\documentclass[10pt,twocolumn,letterpaper]{article}

\usepackage{iccv}
\usepackage{times}
\usepackage{epsfig}
\usepackage{graphicx}
\usepackage{amsmath}
\usepackage{amssymb}
\usepackage{booktabs}
\usepackage{siunitx}
\usepackage{array}
\usepackage{multirow}
\newcolumntype{P}[1]{>{\centering\arraybackslash}p{#1}}
\usepackage{makecell}

\usepackage[pagebackref=true,breaklinks=true,letterpaper=true,colorlinks,bookmarks=false]{hyperref}

\iccvfinalcopy 

\def\iccvPaperID{****} 

\ificcvfinal\fi

\begin{document}

\title{Learning Two-View Correspondences and Geometry \\Using Order-Aware Network}

\author{Jiahui Zhang$^{13}$\thanks{indicates equal contributions.}{ }\thanks{interns at Intel Labs China.} \qquad Dawei Sun$^{2}$\footnotemark[1]{ }\footnotemark[2] \qquad Zixin Luo$^{3}$\thanks{interns at Shenzhen Zhuke Innovation Technology (Altizure).} \qquad Anbang Yao $^{2}$\thanks{indicates corresponding authors.} \qquad Lei Zhou$^{3}$\footnotemark[3] \qquad Tianwei Shen$^{4}$ \\
\qquad Yurong Chen$^{2}$ \qquad Long Quan$^{3}$ \qquad Hongen Liao$^{1}$\footnotemark[4]\\
$^1$Tsinghua University \qquad $^2$Intel Labs China \qquad $^3$Hong Kong University of Science and Technology \\
$^4$Everest Innovation Technology (Altizure) \\
\{jiahui-z15@mails.,liao@\}tsinghua.edu.cn \qquad
\{dawei.sun,anbang.yao,yurong.chen\}@intel.com \\
\{zluoag,lzhouai,quan\}@cse.ust.hk \qquad tianwei@altizure.com
}

\maketitle
\ificcvfinal\thispagestyle{empty}\fi

\begin{abstract}
Establishing correspondences between two images requires both local and global spatial context. Given putative correspondences of feature points in two views, in this paper, we propose Order-Aware Network, which infers the probabilities of correspondences being inliers and regresses the relative pose encoded by the essential matrix. Specifically, this proposed network is built hierarchically and comprises three novel operations. First, to capture the local context of sparse correspondences, the network clusters unordered input correspondences by learning a soft assignment matrix. These clusters are in a canonical order and invariant to input permutations. Next, the clusters are spatially correlated to form the global context of correspondences. After that, the context-encoded clusters are recovered back to the original size through a proposed upsampling operator. We intensively experiment on both outdoor and indoor datasets. The accuracy of the two-view geometry and correspondences are significantly improved over the state-of-the-arts. Code will be available at \url{https://github.com/zjhthu/OANet.git}.

\end{abstract}

\section{Introduction}
Two-view geometry estimation is a fundamental problem in computer vision, which plays an important role in Structure from Motion (SfM) \cite{wu2011visualsfm, schonberger2016structure} and visual Simultaneous Localization and Mapping (SLAM) \cite{mur2015orb}.
Current state-of-the-art SfM \cite{wu2011visualsfm, schonberger2016structure} and SLAM \cite{mur2015orb} pipelines commonly start from local feature extraction and matching.
Outlier rejection algorithm is then applied which is necessary for accurate relative pose estimation.
After that, the relative pose can be recovered from inliers.

\begin{figure}
\centering
\includegraphics[width=0.45\textwidth]{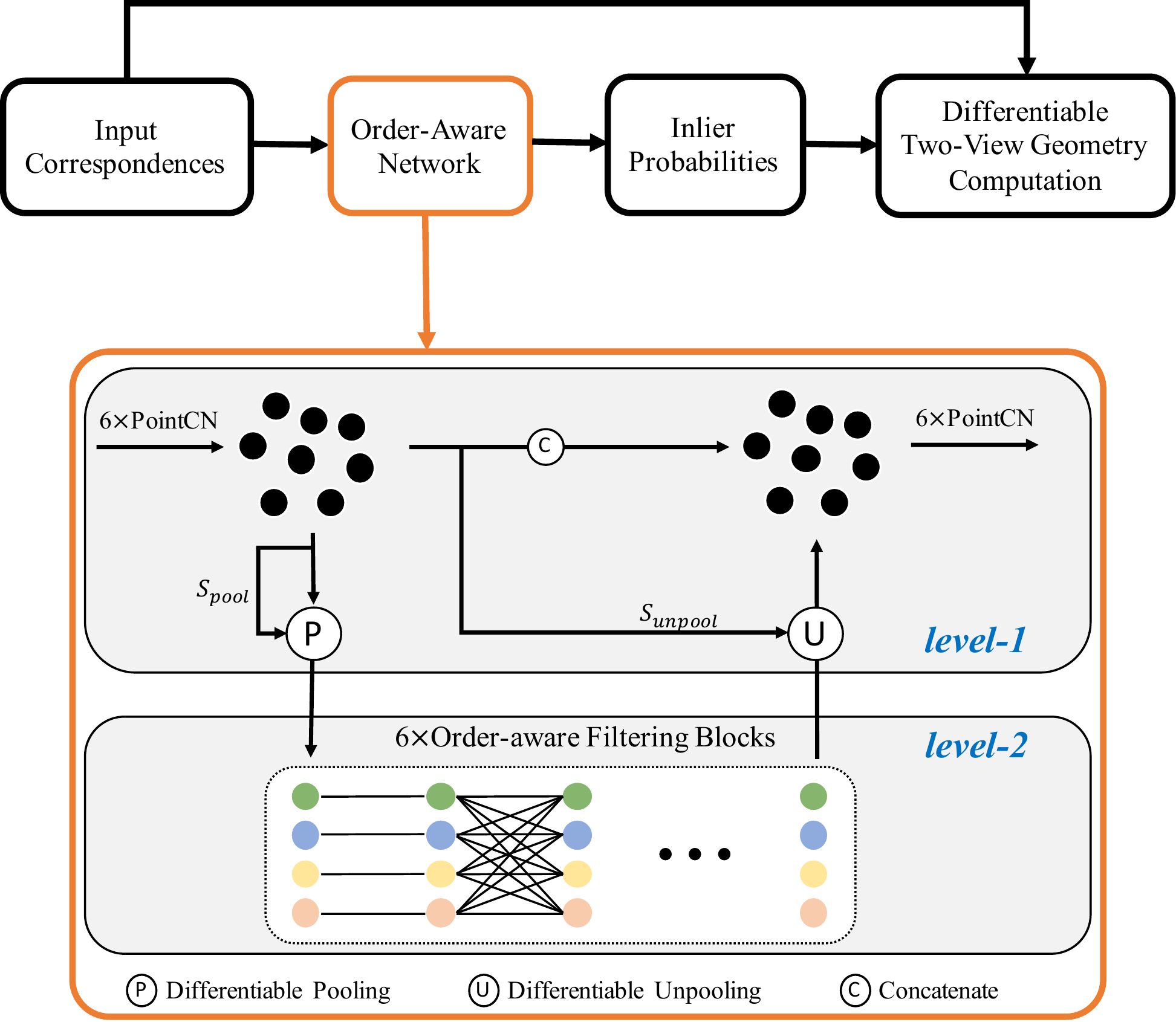}
\caption{The Order-Aware Network to learn two-view correspondences and geometry.
PointCN blocks are used in level-1 to process unordered input.
Besides, we introduce
three novel operations to exploit the local and global context:
(1) The DiffPool layer (left), which
maps unordered nodes to a set of clusters in a canonical order
to caputre local context;
(2) the Order-Aware DiffUnpool layer (right),
which upsamples the clusters using the spatial information of input nodes to build a hierarchical architecture;
(3) the Order-Aware Filtering block in level-2,
which correlates the clusters
thus allows the network to better model the global context.
}
\label{fig:architecture}
\end{figure}

Until recently, great efforts have been spent on applying deep learning techniques to geometric matching pipeline, and most of them focus on learning local feature detectors and descriptors \cite{yi2016lift, detone2017superpoint}. More interestingly, learning-based outlier rejection
has also been revisited \cite{moo2018learning, ranftl2018deep} and achieves appealing results.
Our work also applies learning-based outlier rejection as the core component for two-view geometry estimation.
We exploit a neural network to infer the probability of each correspondence as an inlier, then recover the relative pose by regressing the essential matrix through a closed-form and differentiable computation. The overview of the workflow is illustrated in Fig. \ref{fig:architecture}.





Previous works \cite{moo2018learning,ranftl2018deep} exploited PointNet-like architecture \cite{qi2017pointnet} and Context Normalization \cite{moo2018learning,Dmitry2016Instance} to
classify putative correspondences,
which we refer to as PointCN. It has following drawbacks:
(1) PointNet-like architecture applies Multi Layer Perceptrons (MLPs) on each point individually. Hence it cannot capture the local context \cite{qi2017pointnet++}, \emph{e.g.}, similar motion shared by neighboring pixels \cite{bian2017gms},
which
has been shown to be beneficial for outlier rejection \cite{bian2017gms, zhou2018learning}.
(2) PointCN relies on Context Normalization to encode the global context.
Such a simple operation normalizes the feature maps by their mean and variance, which
overlooks the underlying complex relations among different points and may hinder the overall performance.

One of the challenges in mitigating the above limitations is exploiting neighbors to encoding local context. Unlike 3D point clouds, sparse matches have no
well-defined neighbors,
where this issue is previously tackled in
bilateral domain \cite{lin2014bilateral} (2D spatial domain and 2D motion domain) or by a graphical model \cite{zhou2018learning}.
Besides, another challenge is
modeling the relation between correspondences since they are unordered and have no stable relations to be captured.

To address the above two problems, we draw inspiration from hierarchical representations of Graph Neural Network (GNN).
In particular, we generalize the Differentiable Pooling (DiffPool) \cite{ying2018hierarchical} operator, which is permutation-invariant and originally designed for GNN, into a PointNet-like framework to capture the local context.
Specifically, as shown in Fig. \ref{fig:architecture},
DiffPool maps input nodes to a set of clusters by learning a soft assignment matrix,
instead of using pre-defined heuristic neighbors.
Meanwhile, the permutation-invariant DiffPool essentially yields a canonical order for the resulting clusters, which eschews the need of heuristic sorting such as \cite{niepert2016learning,zhang2018end}.
Moreover, being in a canonical order further enables us to exploit the cluster relation with effective spatially-correlated operators, \emph{i.e.} the proposed Order-Aware Filtering block, to capture more complex global context.
Finally, to assign per-correspondence predictions,
we develop a novel Differentiable Unpooling (DiffUnpool) layer to upsample these clusters to the original size.
It is noteworthy that the proposed DiffUnpool operator is specially designed to be order-aware so as to precisely align the upsampled features with the original input correspondences.

The proposed method is extensively evaluated on both large-scale indoor and outdoor datasets with diverse scenes and achieves significant accuracy improvements on relative pose estimation over the-state-of-the-arts.

Our main contributions are threefold:
\begin{itemize}
\setlength\itemsep{0em}
  \item We introduce the DiffPool and DiffUnpool layers to capture the local context of unordered sparse correspondences in a learnable manner.
  \item
  By the collaborative use of DiffPool operator, we propose Order-Aware Filtering block which exploits the complex global context of sparse correspondences.
  \item Our work significantly improves the relative pose estimation accuracy on both outdoor and indoor datasets.
\end{itemize}

\section{Related Work}
\subsection{Learning based Matching}
With the emergence of deep learning, many works attempted to employ learning-based methods to solve geometric matching tasks, including both dense methods \cite{ummenhofer2017demon, zhou2017unsupervised, li2018undeepvo, rocco2018convolutional} and sparse methods \cite{yi2016lift, ono2018lf, detone2017superpoint, detone2018self, Luo_2018_ECCV, luo2019contextdesc}. For these sparse methods, most of them focused on interest point extraction and description with convolutional neural network (CNN) to replace handcrafted features such as SIFT \cite{lowe2004sift}.
Meanwhile, some works \cite{brachmann2017dsac, moo2018learning, ranftl2018deep} also attempted to solve the outlier rejection problem with learning-based methods to improve the accuracy of relative pose estimation, which is the topic of this work.

\subsection{Outlier Rejection}
Typically, putative correspondences established by handcrafted or learned features contain many outliers, \eg in the wide baseline case. So outlier rejection is necessary to improve relative pose estimation accuracy. RANSAC \cite{fischler1981Ransac} is the standard and still the most popular outlier rejection method. USAC \cite{raguram2013usac}
provided a universal framework for RANSAC variants. BF \cite{lin2014bilateral} utilized a piecewise smoothness constraint on the bilateral domain to filter outliers. GMS \cite{bian2017gms} simplified the idea of smoothness constraints as a statistical formulation.
RMBP \cite{zhou2018learning} defined a graphical model which describes the spatial organization of matches to reject outliers.

In the deep learning era, DSAC \cite{brachmann2017dsac} mimicked the behavior of RANSAC and proposed a differentiable counterpart using probabilistic selection. PointCN \cite{moo2018learning} reformulated the outlier rejection task as an inlier/outlier classification problem and an essential matrix regression problem. It exploited PointNet-like architecture to label input correspondences as either inliers or outliers and introduced a weighted eight-point algorithm to directly regress essential matrix. Context Normalization was proposed which can drastically improve the performance. A concurrent work DFE \cite{ranftl2018deep} also used PointNet-like architecture and Context Normalization
but adopted a different loss function and an iterative network. N$^3$Net \cite{plotz2018neural} inserted soft $k$-nearest neighbors (KNN) layer to augment PointCN. Our work is also built on PointCN but puts effort on improving the local and global contexts
by borrowing ideas from Geometric Deep Learning.

\subsection{Geometric Deep Learning}
Geometric Deep Learning deals with data on non-Euclidean domains, such as graphs \cite{kipf2016semi, monti2017geometric, hamilton2017inductive, fey2018splinecnn} and manifolds \cite{qi2017pointnet, xu2018spidercnn, klokov2017escape, graham20183d, zhang2018efficient}.
PointNet-like architecture
can be regarded as a special case of Graph Neural Network which processes graphs without edges.
Different from 3D point clouds, sparse correspondences have no well-defined neighbors. This is also a difficulty faced by many tasks on graphs \cite{ying2018hierarchical}. Instead of defining heuristic neighbors for correspondences as done in previous works \cite{lin2014bilateral,zhou2018learning}, we exploit Differentiable Pooling \cite{ying2018hierarchical} to cluster nodes in a learnable manner and capture the local context.
However, the original DiffPool Network is not applicable in our case because it
does not give a full size prediction.
Hence, we propose a novel DiffUnpool layer to upsample the coarsened feature maps and build a hierarchical architecture. Moreover, we introduce an Order-Aware Filtering block with spatial connections to capture the global context.

\section{Order-Aware Network}
We will present Order-Aware Network for learning two-view correspondences and geometry, which contains three novel operations: Differentiable Pooling layer, Order-Aware Differentiable Unpooling layer, and Order-Aware Filtering block. The formulation of our problem is first introduced, and then these submodules successively.

\subsection{Problem Formulation}
Given image pairs, the goal of our task is to remove outliers from putative correspondences and recover the relative pose.
More specifically, after extracting keypoints and their descriptors in each image using handcrafted features \cite{lowe2004sift,rublee2011orb} or learned features \cite{yi2016lift, detone2017superpoint},
putative correspondences can be established by finding their nearest neighbors in the other image.
Then outlier rejection method is applied to establish geometrically consistent correspondences. Finally, an essential matrix can be recovered from the inlier correspondences by a closed-form solution \cite{longuet1981EightPoint,moo2018learning}.

The input to the outlier rejection process is a set of putative correspondences:
\begin{equation}
\begin{aligned}
\mathbf{C} = [c_1; c_2; ... ; c_N] \in \mathcal{R}^{N\times4},
c_i = (x_1^i, y_1^i, x_2^i, y_2^i),
\end{aligned}
\end{equation}
where $c_i$ is a correspondence and $(x_1^i, y_1^i)$, $(x_2^i, y_2^i)$ are the coordinates of keypoints in these two images. The coordinates are normalized using camera intrinsics \cite{moo2018learning}.

Following \cite{moo2018learning}, we formulate the two-view geometry estimation task as an inlier/outlier classification problem and an essential matrix regression problem. We use a neural network to predict the probability of each correspondence to be an inlier and then apply the weighted eight-point algorithm \cite{moo2018learning} to directly regress the essential matrix. The architecture can be written as:
\begin{equation}
\mathbf{z} = f_\phi(\mathbf{C}),
\end{equation}
\begin{equation}
\mathbf{w} = \text{tanh}(\text{ReLU}(\mathbf{z})),
\end{equation}
\begin{equation}
\label{eq:weighted8}
\mathbf{\hat E} = g(\mathbf{w,C}),
\end{equation}

where
$\mathbf{z}$ is the logit values for classification. $f_\phi(\cdotp)$ is a permutation-equivariant neural network and $\phi$ denotes the network parameters.
$\mathbf{w}$ is the weights of correspondences. For each weight $w_i \in [0,1)$, $w_i = 0$ means an outlier.
$\text{tanh}$ and ReLU are applied to easily remove outliers \cite{moo2018learning}.
$g(\cdotp,\cdotp)$ in Eq. \ref{eq:weighted8} is the weighted eight-point algorithm. $\mathbf{\hat E}$ is the regressed essential matrix.
$g(\cdotp,\cdotp)$ takes more than eight correspondences and their weights to compute essential matrix via self-adjoint eigendecomposition.
The weighted eight-point algorithm can be more robust to outliers than traditional eight-point algorithm \cite{hartley2003multiple} because it has considered the contribution of each correspondence. Besides, it is differentiable with respect to $\mathbf{w}$ which makes it possible to regress the essential matrix in an end-to-end manner.

The optimization objective of this neural network is to minimize a classification loss and an essential matrix loss as follows:
\begin{equation}
loss = l_{cls}(\mathbf{z, s}) + \alpha l_{ess}(\mathbf{\hat E, E}),
\end{equation}
where
$l_{ess}$ is the essential matrix loss
between the predicted essential matrix $\mathbf{\hat E} $ and the ground truth essential matrix $\mathbf{E}$. It can be a $L2$ loss \cite{moo2018learning}
\begin{equation}
loss_{L2} = \text{min}\{\|\mathbf{\hat E \pm E}\|\}
\end{equation}
or a geometry loss \cite{ranftl2018deep, hartley2003multiple}
\begin{equation}
loss_{geo} = \frac{\mathbf{{(p_2^T \hat E p_1)}}^2} {\mathbf{{\|Ep_1\|}}_{[1]}^2 + \mathbf{{\|Ep_1\|}}_{[2]}^2 + \mathbf{{\|E^{T}p_2\|}}_{[1]}^2 + \mathbf{{\|E^{T}p_2\|}}_{[2]}^2},
\end{equation}
where $\mathbf{p_1, p_2}$ are correspondences and $\mathbf{t}_{[i]}$ denotes the $i$th element of vector $\mathbf{t}$.
$l_{cls}$ is a binary cross entropy loss for the classification term. $\mathbf{s}$ denotes weakly supervised labels for correspondences, which are also derived using the above geometric error, and a threshold of $10^{-4}$ is used to determine valid correspondences.
$\alpha$ is the weight to balance these two losses.

\begin{figure}[t]
\centering
\includegraphics[width=0.4\textwidth]{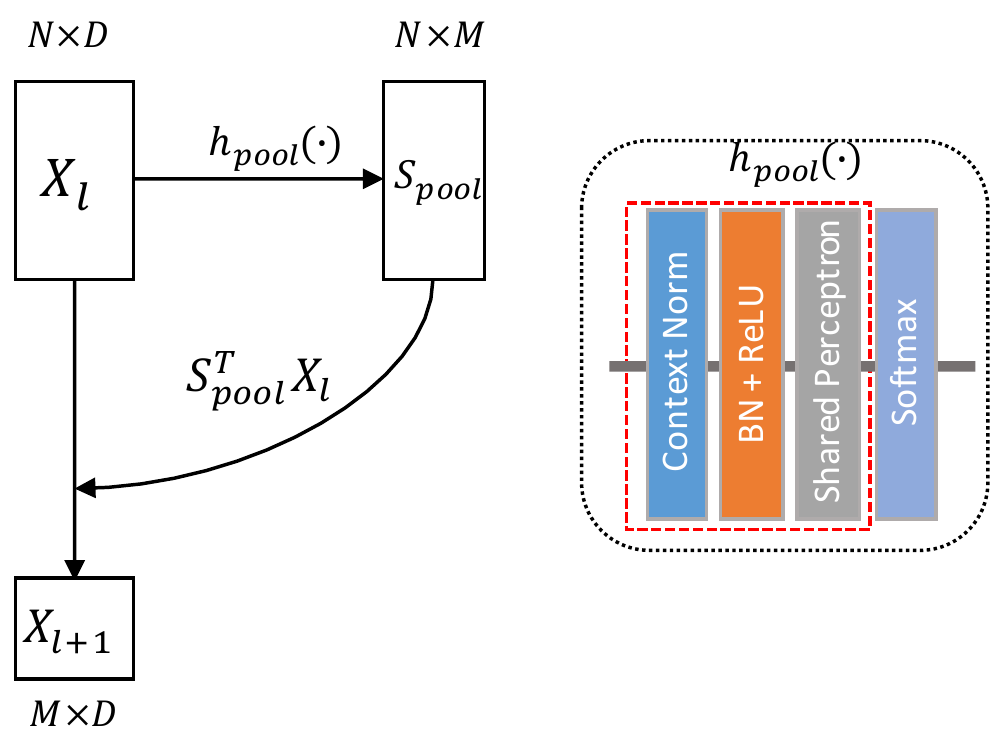}
\caption{Differentiable Pooling layer. DiffPool maps nodes to clusters in a soft assignment manner.
The soft assignment matrix is learned by $h_{pool}(\cdotp)$ which contains one PointCN block (in dashed red box) and one softmax layer.
}
\label{fig:pool}
\end{figure}

\subsection{Differentiable Pooling Layer}
The unordered input correspondences require network $f_\phi(\cdotp)$ to be permutation-equivariant.
So PointNet-like architecture was used \cite{moo2018learning,ranftl2018deep}.
Each block in the PointNet-like \cite{moo2018learning} architecture comprises one Context Normalization layer, one Batch Normalization layer with ReLU, and one
shared Perceptron layer. This so called PointCN block is shown in Fig. \ref{fig:pool}. The proposed Context Normalization layer \cite{moo2018learning} normalizes features of each sample using their statistics and can largely boost the performance.

However, PointNet-like architecture has the drawback in capturing the local context because there is no direct interaction between points.
In order to capture the local context for sparse correspondences, we draw the idea from DiffPool layer \cite{ying2018hierarchical} to learn to cluster nodes to a coarser representation, as shown in Fig. \ref{fig:pool}. The DiffPool layer is analogous to Pooling layer in CNN which assigns nodes to different clusters.
Rather than employing a hard assignment for each node, the DiffPool layer learns a soft assignment matrix.
Denoting the assignment matrix as $\mathbf{S}_{pool} \in \mathcal{R}^{N \times M}$, DiffPool layer maps $N$ nodes to $M$ clusters:

\begin{equation}
\label{equ:assign}
\mathbf{X}_{l+1} = \mathbf{S}_{pool}^T \mathbf{X}_l ,
\end{equation}
where $\mathbf{X}_l \in \mathcal{R}^{N \times D}$ and $\mathbf{X}_{l+1} \in \mathcal{R}^{M \times D}$ are the features at level $l$ and level $l+1$ respectively. $D$ is the dimension of features, and typically $M < N$, \eg $N=2000, M=500$.

As we have mentioned before, the assignment matrix is learned rather than pre-defined. More specifically, taking the features at level $l$, we directly generate the assignment matrix using a permutation-equivariant network as follows:
\begin{equation}
\label{equ:pool}
\mathbf{S}_{pool} = \text{softmax}(h_{pool}(\mathbf{X}_l)),
\end{equation}
where
the permutation-equivariant function $h_{pool}(\cdotp)$ is one PointCN block here.
It maps features from $N\times D$ to $N\times M$. Softmax layer is applied to normalize the assignment matrix along the row dimension. These clusters can be viewed as weighted average results of nodes in the previous level.

\textbf{Permutation-invariance.} DiffPool is a permutation-invariant\footnote{Equivariance means applying a transformation to input equals to applying the same transformation to output, while invariance means applying a transformation to input will not affect the output.} operation \cite{ying2018hierarchical}, which will play a crucial role in our design.
Assuming permuting $\mathbf{X}_{l}$ with a permutation matrix $\mathbf{P} \in \{0,1\}^{N \times N}$, Eq. \ref{equ:pool} becomes
\begin{equation}
\label{equ:pool_permute}
\mathbf{\tilde S}_{pool} = \text{softmax}(h_{pool}(\mathbf{P} \mathbf{X}_l)) = \mathbf{P} \mathbf{S}_{pool},
\end{equation}
because both $h_{pool}(\cdotp)$ and $\text{softmax}$ are permutation-equivariant functions.
So, according to Eq. \ref{equ:assign}, features at level $l+1$ become
\begin{equation}
\label{equ:assign_permute}
\mathbf{X}_{l+1} = \mathbf{\tilde S}_{pool}^T \mathbf{P} \mathbf{X}_l = \mathbf{S}_{pool}^T \mathbf{P}^T \mathbf{P} \mathbf{X}_l = \mathbf{S}_{pool}^T \mathbf{X}_l,
\end{equation}
since $\mathbf{P}^T \mathbf{P} = \mathbf{I}$ holds for every permutation matrix.
Eq. \ref{equ:assign_permute} and Eq. \ref{equ:assign} prove the permutation-invariance property of DiffPool layer.

The permutation-invariance property also means that, once the network is learned, no matter how the input are permuted, they will be mapped into clusters in a particular \textbf{\textit{learned canonical order}} by the DiffPool layer. This canonical order is determined by the parameters of $h_{pool}(\cdotp)$.

\begin{figure}[t]
\centering
\includegraphics[width=0.45\textwidth]{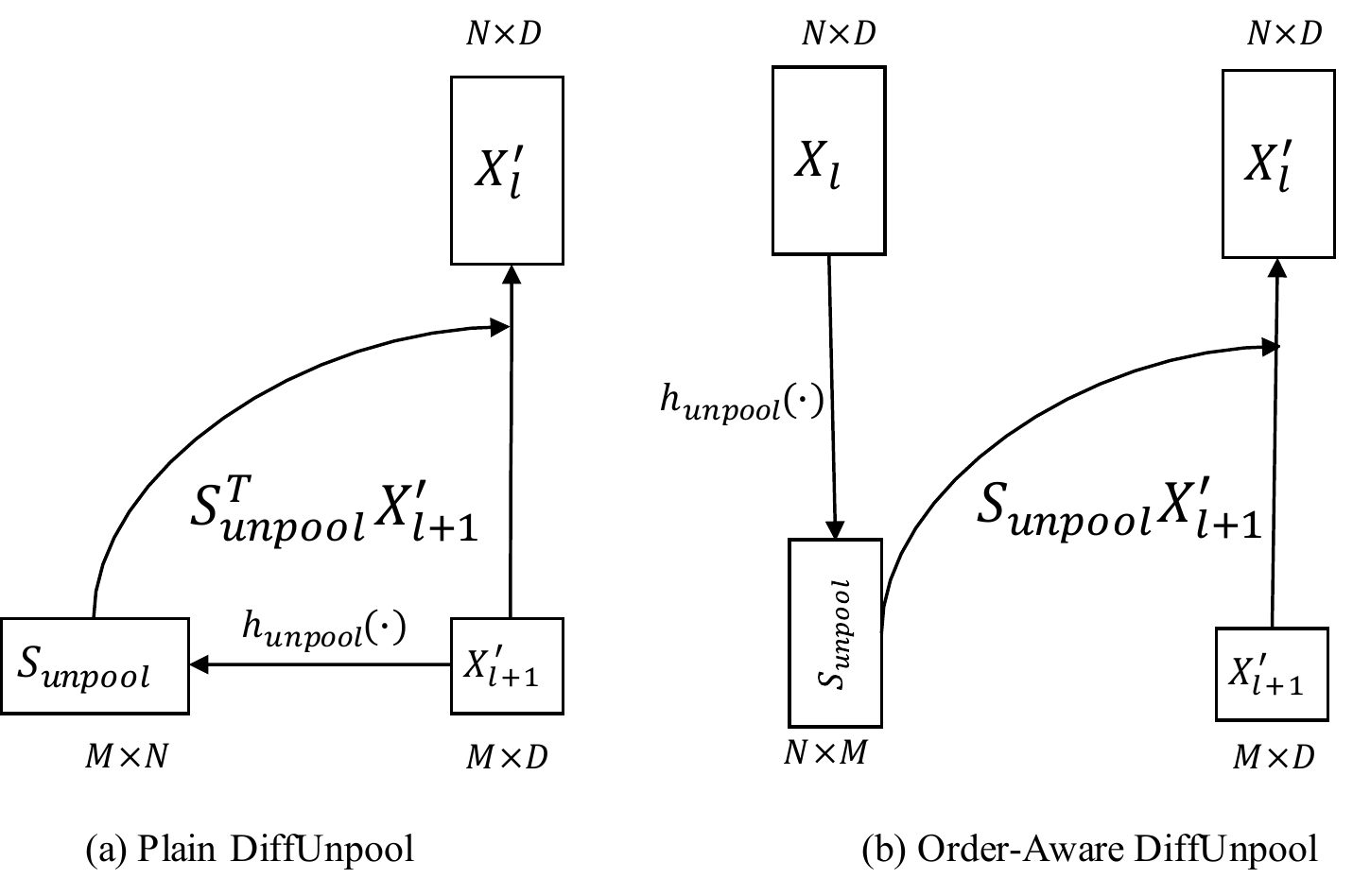}
\caption{Designs of Differentiable Unpooling layer. (a) Plain DiffUnpool layer. It learns a soft assignment matrix using features at level $l+1$. (b) Order-Aware DiffUnpool layer. It learns a soft assignment matrix using features at level $l$ which can encode the order information of nodes at level $l$.
}
\vskip -0.1 in
\label{fig:unpool}
\end{figure}

\subsection{Differentiable Unpooling Layer}
\label{sec:unpool}
DiffPool Network was used to predict the label for an entire graph \cite{ying2018hierarchical}. However, it is not applicable for our sparse matching problem, since we need to give predictions for all correspondences. So we develop a Differentiable Unpooling layer inspired by the DiffPool layer to upsample the coarse representation and build a hierarchical architecture.

A straightforward way to implement the DiffUnpool layer is reversing the behavior of DiffPool layer, as shown in Fig. \ref{fig:unpool}a.
More specifically, similar to Eq. \ref{equ:assign} and Eq. \ref{equ:pool}, an unpooling assignment matrix $\mathbf{S}_{unpool} \in \mathcal{R}^{M \times N}$ is first predicted taking features $\mathbf{X}_{l+1}^{'}$ through:
\begin{equation}
\label{equ:false_unpool_assign}
\mathbf{S}_{unpool} = \text{softmax}(h_{unpool}(\mathbf{X}_{l+1}^{'})),
\end{equation}
where $\mathbf{X}_{l+1}^{'}\in \mathcal{R}^{M \times D}$ denotes new features at the same level of $\mathbf{X}_{l+1}$, and it is
computed from $\mathbf{X}_{l+1}$. We then map features $\mathbf{X}_{l+1}^{'}$ to a new embedding $\mathbf{X}_{l}^{'}\in \mathcal{R}^{N \times D}$ at level $l$ as follows:
\begin{equation}
\mathbf{X}_{l}^{'} = \mathbf{S}_{unpool}^T \mathbf{X}_{l+1}^{'}.
\end{equation}

However, we find the above implementation is not optimal because it cannot align the unpooled features $\mathbf{X}_{l}^{'}$ with features $\mathbf{X}_{l}$ in the previous stage (see section \ref{sec:ablation}).
The point is that DiffPool is a permutation-invariant operation, which means one $\mathbf{X}_{l+1}$ can correspond to various input $\mathbf{X}_l$. In the other words, features $\mathbf{X}_{l+1}$ and $\mathbf{X}_{l+1}^{'}$ at level $l+1$ have lost the spatial order information of features $\mathbf{X}_{l}$ at level $l$.
We cannot expect the learned assignment matrix as in Eq. \ref{equ:false_unpool_assign} can recover the original spatial order of $\mathbf{X}_l$ or generate features which can be precisely aligned with $\mathbf{X}_{l}$, since $\mathbf{S}_{unpool}$ in Eq. \ref{equ:false_unpool_assign} only utilizes information at level $l+1$.

Keeping this in mind, we propose an Order-Aware DiffUnpool layer as shown in Fig. \ref{fig:unpool}b, which can be aware of the particular order (position) of nodes in the previous level.
Different from the above implementation, the assignment matrix for unpooling is learned from features at level $l$ which has stored the input order information as follows:
\begin{equation}
\label{equ:oder_unpool}
\mathbf{S}_{unpool} = \text{softmax}(h_{unpool}(\mathbf{X}_{l})).
\end{equation}
With this unpooling assignment matrix $\mathbf{S}_{unpool} \in \mathcal{R}^{N\times M}$, we can map features at level $l+1$ to level $l$ by:
\begin{equation}
\label{equ:assign_unpool_true}
\mathbf{X}_{l}^{'} = \mathbf{S}_{unpool} \mathbf{X}_{l+1}^{'}.
\end{equation}
Since each row in this $\mathbf{S}_{unpool} \in \mathcal{R}^{N\times M}$ corresponds to one node in $\mathbf{X}_l$, it has already encoded the particular order information of $\mathbf{X}_l$ and ensures the unpooled features can well aligned to the previous stage.
The mapping in Eq. \ref{equ:assign_unpool_true} also requires the learned assignment matrix to be aware of the order of $\mathbf{X}_{l+1}^{'}$. But it is much easier for the network this time since the feature $\mathbf{X}_{l+1}^{'}$ is in a canonical order.
$h_{unpool}(\cdotp)$ in Eq. \ref{equ:oder_unpool} is also a PointCN block and it maps features from $N\times D$ to $N\times M$.
We apply the softmax along the column dimension this time\footnote{Actually we find changing the normalization directions in Eq. \ref{equ:pool} and Eq. \ref{equ:oder_unpool} only has little influence on results. They do not even need to be orthogonal.}, so the unpooled features can be viewed as weighted average results of different clusters.
$\mathbf{X}_{l}^{'}$ is then concatenated with $\mathbf{X}_{l}$ to fuse shallow features.

Another advantage of the proposed Order-Aware DiffUnpool layer is that it does not require a fixed size input.
When there are less than or more than $2000$ keypoints in images, we can still pool nodes to fixed $500$ clusters and then upsample clusters back to the same size. This is useful in practice.
\subsection{Order-Aware Filtering Block}
\label{sec:FC}

With the DiffPool and DiffUnpool layers, we can build a multiscale network which is a common practice in CNN.
We can apply PointCN blocks repeatedly to process these newly generated clusters. However, as we have discussed above, PointCN may have weakness in modeling the complex global context because it ignores the relation between nodes.
Here we propose a simple but more effective operation than PointCN block, which is called Spatial Correlation layer to explicitly model relation between different nodes and capture the complex global context.

\begin{figure}[t]
\centering
\includegraphics[width=0.45\textwidth]{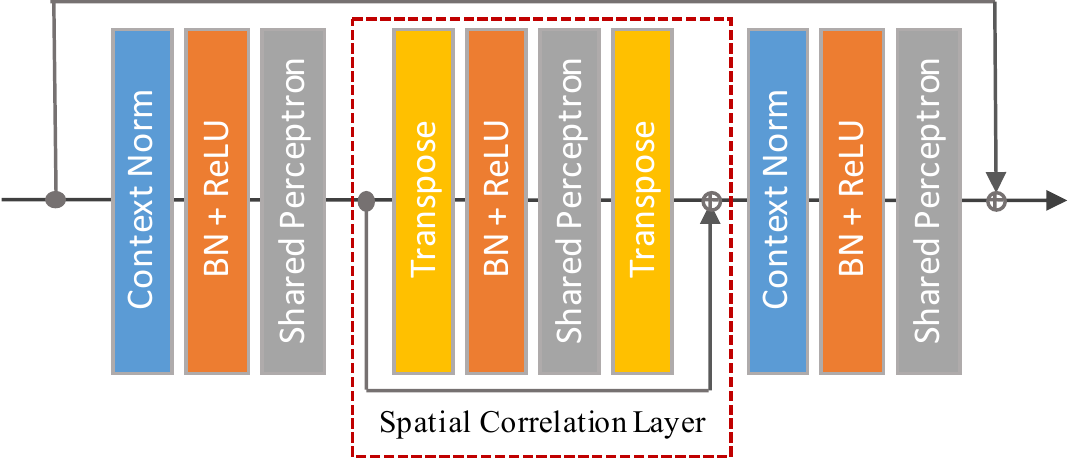}
\caption{Order-Aware Filtering block. We insert the Spatial Correlation layer to PointCN ResNet block. This layer is complementary to PointCN and can help capture the global context effectively.}
\label{fig:oab}
\end{figure}

As we have shown above, {the pooled features are in a canonical order} after the DiffPool layer.
This is a useful property but PointNet-like architecture cannot make full use of it. Our Spatial Correlation layer
applies weight-sharing perceptrons directly on the spatial dimension to establish connections between nodes.
Note this operation is different from the fully connected layer because the weights are shared along the channel dimension, which can help to prevent overfitting.
The Spatial Correlation layer is orthogonal to PointCN, since one is along the spatial dimension and the other is along the channel dimension.
These two operations are complementary, so we assemble them into one block to better capture the global context as shown in Fig. \ref{fig:oab}.

Spatial Correlation layer is implemented by transposing the spatial and channel dimensions of features. After the weight-sharing perceptrons layer, we transpose features back.
Residual connection and batch normalization with ReLU are also used.
We insert the Spatial Correlation layer to the middle of PointCN ResNet block and call this composite module Order-Aware Filtering block which can process data in a canonical order.
Note that before the DiffPool layer, we cannot apply the Spatial Correlation layer on the feature maps as the input data is unordered and there is no stable spatial relation to be captured.
So we apply this simple block only at the level after the DiffPool layer and find it can significantly boost the performance.


\section{Experiments}
We conducted experiments on outdoor YFCC100M \cite{thomee2015yfcc100m} dataset and indoor SUN3D \cite{xiao2013sun3d} dataset. Experiment results and network interpretation are as follows.

\begin{table}[]
\begin{center}
\small
\begin{tabular}{c|cc|ccc}
\Xhline{2\arrayrulewidth}
threshold              & S & L         & mAP5(\%)        & mAP10(\%)       & mAP20(\%)       \\ \Xhline{2\arrayrulewidth}
0.01                   & \checkmark &           & 17.53/12.50 & 27.61/21.15 & 42.06/34.21 \\ \hline
\multirow{2}{*}{0.001} & \checkmark &  & 44.50/12.50 & 54.50/21.15 & 65.27/34.21 \\
                       &   & \checkmark         & 47.98/23.55 & 58.13/36.58 & 68.67/53.08 \\ \Xhline{2\arrayrulewidth}
\end{tabular}
\end{center}
\caption{
Performances of baseline network \cite{moo2018learning} on YFCC100M unknown sequences.
Results \textbf{with/without} RANSAC under error thresholds of \ang{5}, \ang{10} and \ang{20} are all reported.
Changing the inlier threshold in RANSAC and using more data can significantly boost the performance. \textbf{S}: using only sequences {`Saint Peter{'}s'} and {`{brown\_bm\_3\_05}'} as \cite{moo2018learning}. \textbf{L}: using 68 sequences.
}
\label{tab:baseline}
\end{table}

\subsection{Datasets}
\label{sec:datasets}

\textbf{Outdoor scenes.} We use the Yahoo{'}s YFCC100M dataset \cite{thomee2015yfcc100m}, which contains 100 million photos from internet. The authors of \cite{heinly2015reconstructing} later generated 72 3D reconstructions of tourist landmarks from a subset of the collections. We use four sequences \cite{moo2018learning}
as unknown scenes to test generalization ability. For training sequences, different from PointCN,
we use the remaining 68 sequences for training, while \cite{moo2018learning} uses only two sequences. Our setting is not prone to overfitting on known sequences and has better generalization ability as shown in Tab. \ref{tab:baseline}.
To have a fair comparison, we re-train all models on the same data.

Minimum visual overlap is required if pairs are selected into the dataset. For outdoor scenes, the overlap is the number of sparse 3D points in the reconstructed model which can be both seen by the image pairs.
We use the camera poses and sparse models provided by \cite{heinly2015reconstructing} to generate ground-truth.

\textbf{Indoor scenes.} We use the SUN3D dataset \cite{xiao2013sun3d} for indoor scenes, which is an RGBD video dataset with camera poses computed by generalized bundle adjustment. Following \cite{ummenhofer2017demon} we split the dataset into 253 scenes for training and 15 as unknown scenes for testing. This splitting can ensure there is no spatial overlap between training and testing datasets. We find some sequences in the training set do not provide camera poses, so we drop these sequences and finally get 239 sequences for training. We subsample videos every 10 frames. The visual overlap for indoor scenes is computed by projecting the depth map to the other image.

Following \cite{moo2018learning}, we test on both known scenes and unknown scenes. The known scenes are the training sequences. We split them into disjoint subsets for training ($60\%$), validation ($20\%$) and testing ($20\%$). The unknown sequences are the test sequences described above.


\subsection{Evaluation Metrics}
We use the angular differences between ground truth and predicted vectors for both rotation and translation as the error metric. mAP results with and without RANSAC post-processing are reported. We find the inlier threshold of OpenCV function \texttt{findEssentialMat()} used in \cite{moo2018learning} is not optimal. Changing the threshold from $0.01$ to $0.001$ will largely improve results with RANSAC, as shown in Tab. \ref{tab:baseline}. We will use mAP under \ang{5} as the default metric since it is more usable in 3D reconstruction context.

\subsection{Implementation Details}
The baseline network \cite{moo2018learning} has 12 PointCN ResNet blocks. Based on this network, we add one DiffPool layer and one DiffUnpool layer. Another 6 Order-Aware Filtering blocks at the second level are used as shown in Fig. \ref{fig:architecture}. The channel dimensions are all 128 in these blocks.
The inputs to the network are $N\times 4$ putative correspondences established using SIFT feature, typically $N=2000$. After DiffPool layer, the number of nodes are reduced to fixed 500 which gives best performance.
Besides, we also use an iterative network as \cite{ranftl2018deep} which takes residuals and weights of previous stage as additional inputs. This can futher improve the performance.
Our network is implemented with Pytorch \cite{paszke2017automatic}. We use Adam solver with a learning rate of $10^{-4}$ and batch size 32. Weight $\alpha$ is $0$ during the first 20k iterations and then $0.1$ in the rest $480$k iterations as in \cite{moo2018learning}.

\subsection{Ablation Studies}
\label{sec:ablation}
In this section, we will give ablation studies about the proposed operations, loss functions and network architecture on YFCC100M dataset.

\begin{table}[]
\begin{center}
\resizebox{\linewidth}{!}{
\begin{tabular}{p{0.9cm}p{0.5cm}p{0.5cm}p{0.5cm}p{0.5cm}p{0.5cm}p{0.5cm}|cc}
\Xhline{2\arrayrulewidth}
PointCN    & {UnA}   & {UnB}   & OF        & L3    & Geo    & Iter       & Known          & Unknown         \\ \hline
\checkmark &            &            &            &      &  &      & 34.36/13.93 & 47.98/23.55 \\ \hline
\checkmark & \checkmark &            &            &      &  &      & 34.38/14.04 & 47.93/24.10\\ \hline
\checkmark &            & \checkmark &            &      &  &      & 36.33/17.88 & 49.65/28.78 \\ \hline
\checkmark &            & \checkmark & \checkmark &      &  &      & 40.78/25.94 & 51.63/32.55 \\ \hline
\checkmark &            & \checkmark & \checkmark & \checkmark &  & &  39.69/26.04 & 50.70/30.48 \\ \hline
\checkmark &            & \checkmark & \checkmark &  & \checkmark &  & 40.79/28.39 & 51.10/33.68  \\ \hline
\checkmark &            & \checkmark & \checkmark &  & \checkmark & \checkmark & \textbf{42.46}/33.06 & \textbf{52.18}/39.33  \\ \Xhline{2\arrayrulewidth}
\end{tabular}
}
\end{center}
\caption{Ablation study on YFCC100M. mAP (\%) on both known and unknown scenes are reported \textbf{with/without} RANSAC post-processing. \textbf{UnA}: the plain DiffUnpool layer. \textbf{UnB}: the Order-Aware DiffUnpool layer. \textbf{OF}: using the Order-Aware Filtering blocks rather than PointCN blocks in the second level. \textbf{L3}: a larger model with three levels. \textbf{Geo}: using geometry loss rather than $L2$ loss. \textbf{Iter}: using the iterative network.}
\label{tab:ablation}
\end{table}


\begin{figure*}
\centering
\includegraphics[width=\textwidth]{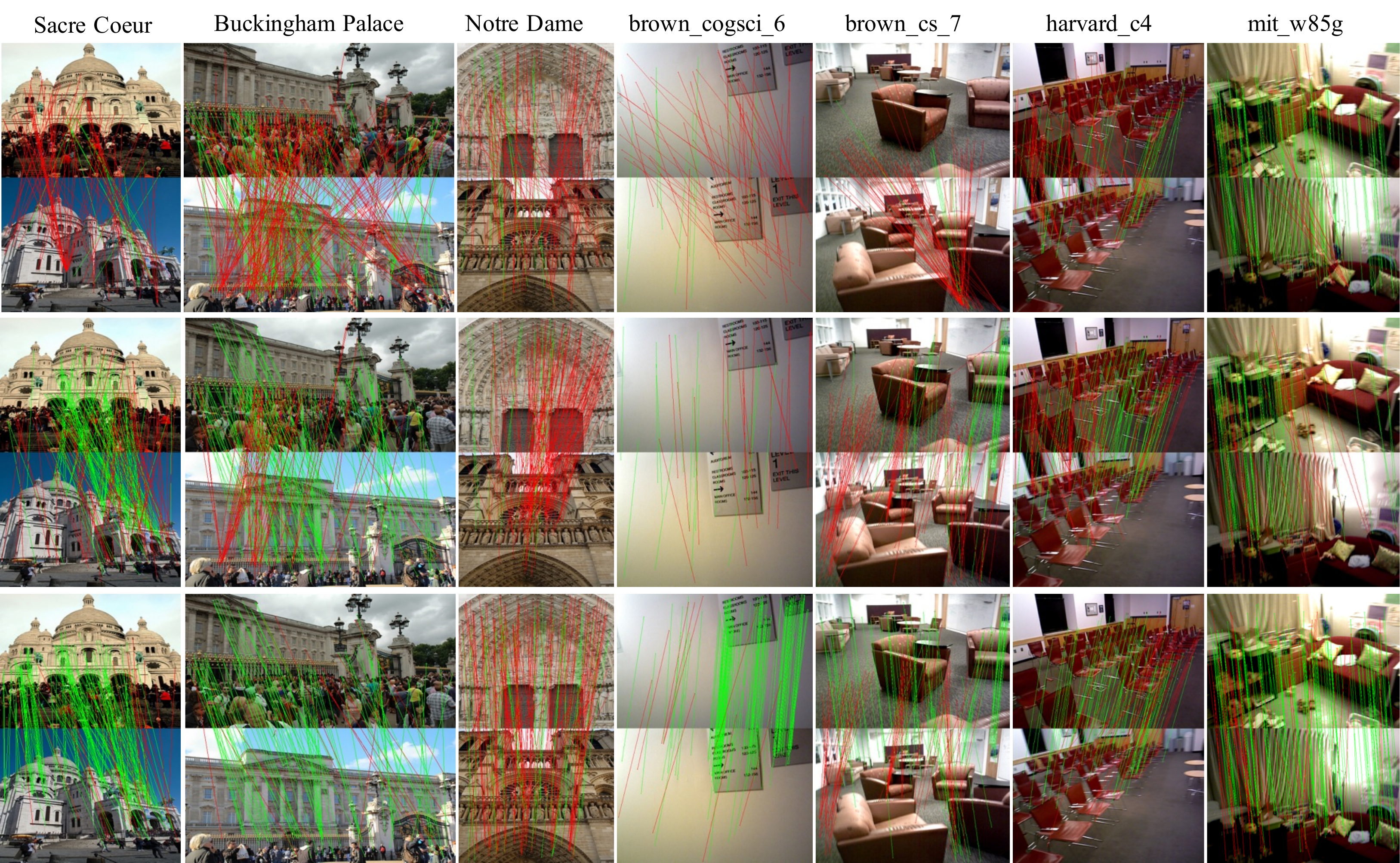}
\caption{Matching results using RANSAC (top), PointCN \cite{moo2018learning} (middle) and our method (bottom). Images are taken from test set of YFCC100M and SUN3D datasets. Correpondences are in green if they conform the ground truth essential matrix (true positives), and in red otherwise (false positives). \textbf{Best viewed in color}.}
\label{fig:result}
\end{figure*}




\textbf{DiffUnpool layer design.} To demonstrate the efficacy of DiffUnpool layer, we add DiffPool and DiffUnpool layers to the baseline PointCN model.
Both plain DiffUnpool and Order-Aware DiffUnpool described in section \ref{sec:unpool} are tested. After the DiffPool layer, another six PointCN ResNet blocks are used.
Features after DiffUnpool layer are concatenated to the previous stage. As shown in Tab. \ref{tab:ablation}, our Order-Aware DiffUnpool (PointCN + UnB) achieves an improvement of 5.23\% over the baseline on unknown scenes when without RANSAC, while the plain DiffPool (PointCN + UnA) gives a negligible improvement over the baseline.

\textbf{Plain PointCN block \emph{vs.} Order-Aware Filtering block.} We replace the PointCN blocks at the second level with Order-Aware Filtering blocks described in section \ref{sec:FC}, which can better exploit the spatial relationships within the clusters. As shown in Tab. \ref{tab:ablation}, the proposed block (PointCN + UnB + OF) can significantly boost the performance over simple PointCN block (PointCN + UnB), achieving an improvement of 3.77\% on unknown scenes without RANSAC.


\textbf{Does a larger model help?} We train a larger model which is a U-Net \cite{ronneberger2015u} with three levels. 12 PointCN ResNet blocks are used at the first level, 12 and 6 Order-Aware Filtering blocks are used at the second and third level. The number of nodes in the second and third level is 500 and 125 respectively. However, we find this larger model even drops on unknown scenes, as shown in Tab. \ref{tab:ablation}. This might show that the representational ability of Order-Aware Filtering block suffices to capture the global context.
So we use the two-level network in our rest experiments.

\textbf{Essential matrix loss.} $L2$ loss is used as the essential matrix loss in previous experiments. However, the $L2$ loss is not geometric meaningful. So we replace the $L2$ loss with the Gold Standard geometry loss \cite{ranftl2018deep, hartley2003multiple}. $\alpha$ is set to 0.5. Clamping the geometry losses to 0.1 works best in our case. Using the geometry loss helps a little for both known and unknown scenes as shown in Tab. \ref{tab:ablation}.

\textbf{Iterative network.} Iterative network shares similarity \cite{ranftl2018deep} with traditional guided matching method. Residuals and weights are passed to next stage iteratively to guide the estimation. Here we use one initialization network and one refinement network. Each network has 6 PointCN ResNet blocks and 3 Order-Aware Filtering blocks to keep almost the same amount of parameters. We find it is really necessary to detach the gradients from latter stage. Tab. \ref{tab:ablation} shows that the iterative network can largely improve the mAP from 33.68\% to 39.33\% without RANSAC on unknown scenes.

\begin{table}[]
\begin{center}
\small
\resizebox{\linewidth}{!}{
\begin{tabular}{c|cc|cc}
\Xhline{2\arrayrulewidth}
        & \multicolumn{2}{c|}{Outdoor(\%)} & \multicolumn{2}{c}{Indoor(\%)} \\ \cline{2-5}
        & Known       & Unknown     & Known       & Unknown     \\ \Xhline{2\arrayrulewidth}
RANSAC  & 5.82/-      & 9.08/-      & 4.38/-      & 2.86/-      \\
PointCN\cite{moo2018learning} & 34.36/13.93 & 47.98/23.55 & 20.44/11.28 & 15.98/9.36  \\
PointNet++\cite{qi2017pointnet}  &34.15/9.28 & 46.23/14.04 & 20.28/7.15 & 15.61/5.59  \\
N$^3$Net\cite{plotz2018neural}& 34.18/12.49 & 49.13/23.18 & 20.31/7.95 & 15.38/7.13 \\
DFE\cite{ranftl2018deep} & 36.87/18.40 & 49.45/29.70 & 20.97/14.09 & 16.45/12.45\\
Ours    & 40.78/25.94 & 51.63/32.55 & 21.82/16.09 & 16.51/12.54 \\
Ours++    & \textbf{42.46}/33.06 & \textbf{52.18}/39.33 & \textbf{22.50}/21.44 & \textbf{17.50}/16.39 \\ \Xhline{2\arrayrulewidth}
RANSAC* & 15.21/-     & 21.95/-     & 18.17/-     & 14.58/-      \\
PointCN*\cite{moo2018learning}& 30.48/13.82 & 43.18/24.83 & 23.66/12.04 & 18.52/10.21  \\
Ours*   & \textbf{33.42}/23.85 & \textbf{46.28}/32.18 & \textbf{24.31}/14.81 & \textbf{19.04}/12.12 \\ \Xhline{2\arrayrulewidth}
\end{tabular}
}
\end{center}
\caption{Comparision with other baselines on YFCC100M and SUN3D. mAP (\%) (\textbf{with/without} RANSAC post-processing) on are reported. \textbf{Ours++} uses the geometry loss and iterative network while \textbf{Ours} not use.
Methods with {*} means using SuperPoint \cite{detone2017superpoint}, otherwise using SIFT.
}
\label{tab:comparison}
\end{table}
\subsection{Comparison to Other Baselines}
We compare our network with other state-of-the-art models from \cite{moo2018learning, plotz2018neural, qi2017pointnet, ranftl2018deep} on both outdoor and indoor datasets.
All these models are trained under the same settings.
For N$^3$Net \cite{plotz2018neural}, we use the official implementation. We find N$^3$Net is unstable during training, so we run it for three times and give the best results here.
PointNet++ \cite{qi2017pointnet} is an extension of PointNet which also aims to improve the capability in capturing local context of point sets. As we have discussed before, it may not be optimal for our sparse matching problem because correspondences have no well-defined neighbors.
Here we implement a 4D-version PointNet++ which exploits the 4D Euclidean space as the underlying metric space.
DFE \cite{ranftl2018deep} is a concurrent work with \cite{moo2018learning} and has similar core designs. We implement \cite{ranftl2018deep} based on \cite{moo2018learning} by adopting their loss formulation and iterative network with the authors{'} help.

Results are shown in Tab. \ref{tab:comparison}, our method achieves best results under all settings, showing improvements of 15.78\% and 7.03\% over PointCN \cite{moo2018learning} on both outdoor and indoor unknown scenes without RANSAC and still works well with strong RANSAC post-processing. We also provide the precision (inlier ratio), recall and F-score of each method in supplementary material.
Fig. \ref{fig:result} shows the visualization results of our method and other baselines. It can be found that our method can give better results on several difficult scenes such as wide baselines, textureless objects, repetitive structures, and large illumination changes.

We also evaluate learned features such as SuperPoint \cite{detone2017superpoint} as shown in Tab. \ref{tab:comparison}.
It is surprising to find SuperPoint gives worse results in outdoor scenes than SIFT when using learned outlier rejection methods. Although it performs much better than SIFT when only using RANSAC.
It might demonstrate that SuperPoint has better descriptors but less accurate keypoints.
It can give putative correspondences with higher inlier ratio thus has better performance when only using RANSAC.
But the bottleneck may become keypoint accuracy when inlier ratio is largely improved, in which situation, SuperPoint performs worse.

\begin{figure}[t]
\centering
\includegraphics[width=0.48\textwidth]{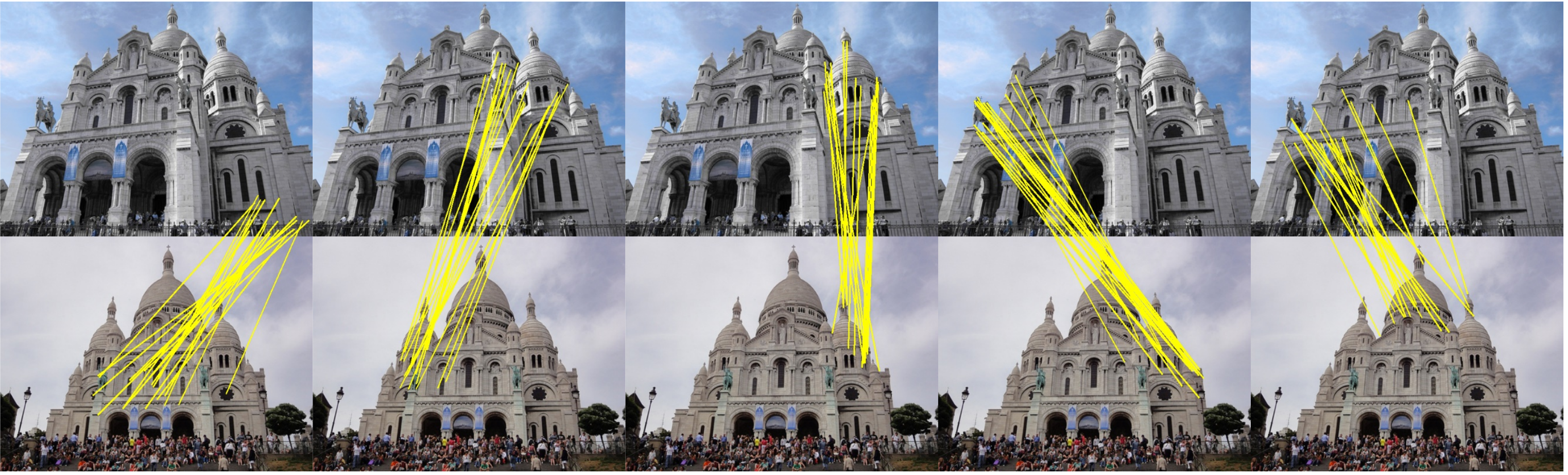}
\caption{DiffUnpool layer visualization. Top 15 responses in different columns of $S_{unpool}$ are visualized in the same image pair. Different clusters might correspond to different motions in different areas.
\textbf{Best viewed in color with \textbf{$200\%$} zoom in.}
}
\label{fig:unpool_channel}
\end{figure}

\begin{figure}[t]
\centering
\includegraphics[width=0.5\textwidth]{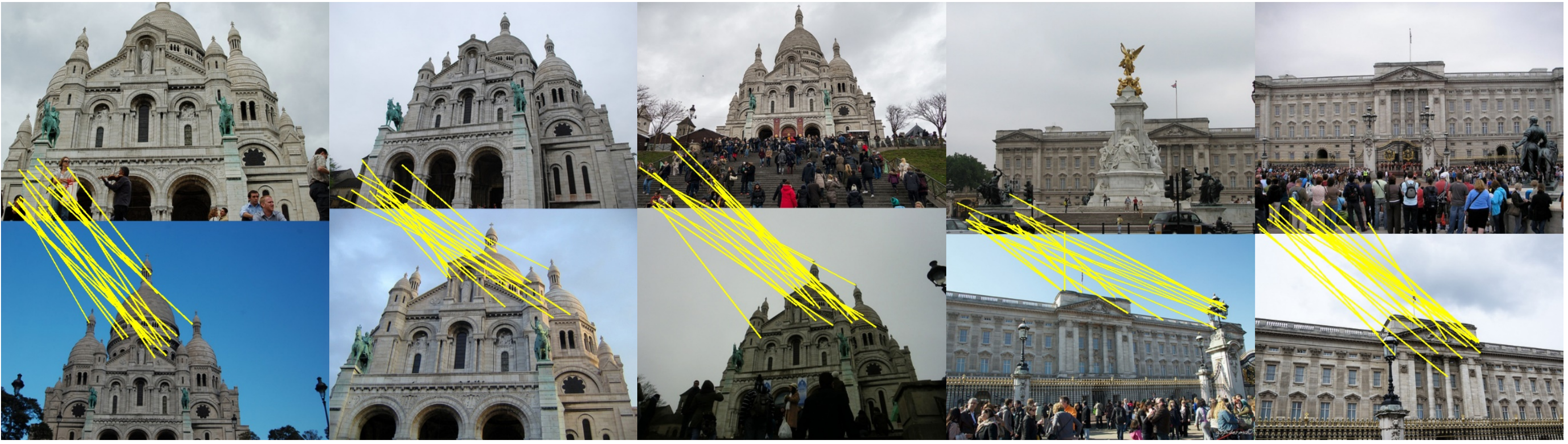}
\caption{DiffUnpool layer visualization. Top 20 responses in the same column of $S_{unpool}$ are visualized in different image pairs. Motions in different pairs are roughly consistent.
\textbf{Best viewed in color with \textbf{$200\%$} zoom in.}
}
\label{fig:unpool_motion}
\end{figure}

\subsection{Network Visualization}
In order to understand the mechanism of the proposed Order-Aware Network, we visualize the assignment matrix $S_{unpool} \in \mathcal{R}^{N\times M}$ of DiffUnpool layer which
reflects the spatial relationships between different nodes in the first level.
More specifically, we visualize the top $k$ responses in each column of $S_{unpool}$.
Each column in $S_{unpool}$ represents one cluster and
each row corresponds to one putative correspondence.
These top $k$ correspondences are ``clustered'' together because they all have a strong response to the same cluster.
We find DiffUnpool can capture meaningful context for sparse matching.
Fig. \ref{fig:unpool_channel} shows that different clusters might correspond to different local motions.
Moreover, we find the corresponding motion of a particular cluster are roughly consistent in different pairs and even in different scenes as shown in Fig. \ref{fig:unpool_motion}, which supports that the pooled features are in a canonical order.

\section{Conclusion}
In this work, we proposed the Order-Aware Network for learning two-view correspondences and geometry. The introduced DiffPool layer and Order-Aware DiffUnpool layer can learn to cluster meaningful nodes to capture local context. Besides, we develop Order-Aware Filtering blocks to capture the global context. These operations can significantly improve relative pose estimation accuracy on both outdoor and indoor datasets.
\section*{Acknowledgment}
This work was supported by
National Natural Science Foundation of China (81427803, 81771940),
National Key Research and Development Program of China (2017YFC0108000),
Beijing Municipal Natural Science Foundation (7172122, L172003)
and Soochow-Tsinghua Innovation Project (2016SZ0206).
Part of work was done when Jiahui Zhang was an intern at ILC and visiting HKUST.
We also thanks Vladlen Koltun, Ren{\'e} Ranftl and David Hafner for helping us reimplement their work.
\clearpage
{\small
\bibliographystyle{ieee_fullname}
\bibliography{egbib}
}

\clearpage
\section*{A. Supplementary appendix}

\subsection*{A.1 Weighted Eight-Point Algorithm}
Here we provide a detailed description of the weighted eight-point algorithm \cite{moo2018learning}.
Given $N$ correspondences $c_i = (x_1^i, y_1^i, x_2^i, y_2^i), 1\leq i \leq N$,
we can construct a matrix $\mathbf{X} \in \mathcal{R}^{N\times 9}$,
where each row has the form of $[x_1^i x_2^i, x_1^i y_2^i, x_1^i, y_1^i x_2^i, y_1^i y_2^i, y_1^i, x_2^i, y_2^i, 1]$.

Traditional eight-point algorithm \cite{hartley2003multiple} seeks to minimize $\| \mathbf{X}^T \mathbf{X} \text{Vec}(\mathbf{E}) \|$ to recover the essential matrix $\mathbf{E}$, where $\text{Vec}(\cdot)$ is the vectorization operation.
The weighted eight-point algorithm extends traditional eight-point algorithm to a weighted formulation, which minimizes $\| \mathbf{X}^T \text{diag}(\mathbf{w}) \mathbf{X} \text{Vec}(\mathbf{E}) \|$, where $\mathbf{w}$ is the probabilities predicted by the neural network.
This linear least square problem has a closed-form solution that $\text{Vec}(\mathbf{E})$ is the eigenvector associated to the smallest eigenvalue of $\| \mathbf{X}^T \text{diag}(\mathbf{w}) \mathbf{X}\|$. The eigendecomposition operation is also differentiable according to \cite{ionescu2015matrix} which makes the essential matrix regression term can be trained end-to-end.

\subsection*{A.2 Outlier rejection results of different methods}
For completeness, we also provide the precision (inlier ratio), recall and F-score of each method in Tab. \ref{tab:inlier}.
\begin{table}[h]
\begin{center}
\resizebox{8cm}{!}{
\begin{tabular}{cccc|ccc}
\Xhline{2\arrayrulewidth}
        & \multicolumn{3}{c|}{Outdoor}                    & \multicolumn{3}{c}{Indoor}                        \\ \hline
        & precision (\%) & recall (\%)    & F-score        & precision (\%) & recall (\%)    & F-score        \\ \hline
RANSAC  & 41.83          & 57.08          & 48.28          & 44.11          & 46.42          & 45.24          \\
PointCN\cite{moo2018learning} & 51.18          & 84.81          & 63.84          & 45.45          & 82.95          & 58.72          \\
Ours    & \textbf{54.55} & \textbf{86.67} & \textbf{66.96} & \textbf{46.95} & \textbf{83.78} & \textbf{60.18} \\ \Xhline{2\arrayrulewidth}
\end{tabular}
}
\end{center}
\caption{Result comparison of different methods. Inlier threshold is $10^{-4}$ of symmetric epipolar distance.
}
\label{tab:inlier}
\end{table}
\end{document}